\documentclass[runningheads]{llncs}
\usepackage[T1]{fontenc}
\usepackage{graphicx}
\usepackage{hyperref}
\usepackage{amsmath,amssymb}
\usepackage{booktabs}
\usepackage[table]{xcolor}
\usepackage{cleveref}
\begin{document}

\title{MAMVI: 3D Test-Time Adaptation via Masked Multi-View Point Clouds}
\titlerunning{MAMVI: 3D Test-Time Adaptation via Masked Multi-View Point Clouds}
%
\author{Inseok Kong\inst{1} \and Geunyoung Jung\inst{2} \and Jiyoung Jung\inst{2, *}}

\authorrunning{I. Kong et al.}

\institute{Department of Geo Informatics, University of Seoul \and
Department of Artificial Intelligence, University of Seoul\\
\email{\{kong7963,gyjung975,jyjung\}@uos.ac.kr}}

\maketitle

\makeatletter
\def\blfootnote{\gdef\@thefnmark{}\@footnotetext}
\makeatother

\blfootnote{\textsuperscript{*} Corresponding author}
\begin{abstract}
3D point cloud models suffer significant performance degradation under distribution shifts caused by sensor noise, occlusions, and environmental changes. Test-time adaptation (TTA) has emerged as a practical paradigm for mitigating this issue during inference. Recently, leveraging multi-view augmentation has shown promise in improving 3D TTA performance. However, existing multi-view approaches are often constrained by sequential optimization that treats each view independently. This sequential optimization leads to substantial inference latency due to repetitive optimization steps, making real-time adaptation impractical. To address this, we propose Masked Multi-View Test-Time Adaptation (MAMVI), which replaces sequential optimization with a unified single-step adaptation. Specifically, MAMVI utilizes a hybrid masking strategy that combines fixed ratios for stability with Beta-distributed sampling for diversity. By aggregating losses across multiple views, MAMVI performs adaptation through a single backward pass based on multi-view consensus. Additionally, a confidence-based adaptive learning rate is used to dynamically adjust the adaptation intensity for each sample. Extensive experiments on ModelNet-40C, ShapeNet-C, and ScanObjectNN-C demonstrate that MAMVI achieves state-of-the-art accuracy on ShapeNet-C and ScanObjectNN-C. Moreover, it remains competitive on ModelNet-40C while delivering 4.9--8.9$\times$ faster inference, making it highly suitable for real-time applications. The code is available at {\url{https://github.com/Inseok-kong/MAMVI}}.

\end{abstract}

\section{Introduction}
\label{sec:intro}

Deep learning has enabled remarkable progress in 3D point cloud classification~\cite{qi2017pointnet,qi2017pointnet++,wang2019dynamic,pang2023masked,zhang2022point}. However, model performance can degrade significantly~\cite{sun2022benchmarking} when the test (target) distribution differs from the training (source) distribution. These distribution shifts in 3D data vary widely due to sensor differences (RGB-D, LiDAR), environmental changes (lighting, weather), and occlusions. Given the vast range of possible distribution shifts, pre-training a model for every potential scenario is impractical. It is thus essential to develop methods that can adapt to these distribution shifts at test time without supervision.

Test-Time Adaptation (TTA) has emerged as a practical paradigm for alleviating performance degradation induced by distribution shifts at inference time.  Without requiring additional labeled data or full retraining, TTA enables pre-trained models to adapt on-the-fly using unlabeled test samples, allowing them to effectively handle previously unseen target distributions during inference.  While TTA has been extensively studied in the 2D domain~\cite{wang2020tent,liang2020we,lee2013pseudo,iwasawa2021test,boudiaf2022parameter,mirza2022norm,zhang2022memo,niu2023towards}, most existing methods are designed for regular grid-structured images and therefore cannot be generalized to sparse and irregular 3D point clouds. Consequently, extending TTA to the 3D domain remains a non-trivial challenge. To address this, recent works have made initial efforts to explore TTA for 3D point cloud classification. Inspired by Test-Time Training (TTT)~\cite{sun2020test,liu2021ttt++,gandelsman2022testtime}, MATE~\cite{mirza2023mate} and SMART-PC~\cite{bahri2025smartpc} incorporate auxiliary self-supervised tasks during the pre-training stage. However, these approaches require modifying network architectures and retraining models with additional losses on source data, which significantly limits their practicality and applicability in real-world applications. CloudFixer~\cite{shim2024cloudfixer} and 3DD-TTA~\cite{dastmalchi2025test} perform input-level TTA by leveraging diffusion models to transform input data to the source distribution. Although they do not involve any modification to the pre-training stage, they still rely on source data and incur substantial computational overhead due to the training and inference of diffusion models. For more efficient approaches, BFTT3D~\cite{wang2024backpropagation} and Purge-Gate~\cite{yazdanpanah2025purge} introduce backpropagation-free adaptation, but their performance relies heavily on feature prototypes extracted from source data.
Recently, model-level TTA has attracted increasing attention as it directly adapts pre-trained models at inference time, thereby resolving the source data dependency. SVWA~\cite{bahri2025test} achieves strong performance through multi-view augmentation and weight averaging. However, it assigns a separate model instance to each augmented view and optimizes them independently via sequential optimization, leading to significant inference latency.

To improve inference efficiency without source data dependency, we propose Masked Multi-View Test-Time Adaptation (MAMVI), which eliminates the latency of sequential optimization by enabling a unified single-step adaptation. Specifically, MAMVI employs a hybrid masking strategy that combines fixed ratios for stability with Beta-distributed sampling for diversity. By aggregating losses across multiple masked observations, our framework facilitates reliable consensus-based adaptation with only one backward pass, independent of the number of views. Furthermore, by generating these views through efficient masking instead of repeating Farthest Point Sampling (FPS), MAMVI significantly improves inference throughput. The framework adapts normalization parameters using confidence-based adaptive learning rates, ensuring effective distribution alignment while preserving inherent source knowledge.

We conduct extensive experiments on three widely-used 3D point cloud corruption benchmarks: ModelNet-40C, ShapeNet-C, and ScanObjectNN-C. The results demonstrate that our approach, MAMVI, achieves competitive performance gains over comparable methods.

Our contributions are as follows:

\begin{itemize}
\renewcommand{\labelitemi}{$\bullet$}
\item We introduce MAMVI, which generates multiple masked views using a hybrid sampling strategy that incorporates a Beta distribution.
\item We enable efficient multi-view TTA by performing a single-step adaptation with a unified backward pass on a single model weight.
\item MAMVI achieves state-of-the-art accuracy on ScanObjectNN-C and ShapeNet-C and competitive results on ModelNet-40C, while delivering 4.9--8.9$\times$ faster inference speed compared to prior multi-view methods.
\end{itemize}

\section{Related Work}

\subsection{Test-Time Adaptation}

Test-Time Adaptation (TTA) has been extensively studied in 2D vision through various approaches including entropy minimization~\cite{wang2020tent,liang2020we,niu2023towards}, pseudo-labeling~\cite{lee2013pseudo}, test-time augmentation~\cite{zhang2022memo}, classifier adjustment~\cite{iwasawa2021test}, Laplacian regularization~\cite{boudiaf2022parameter}, batch normalization calibration~\cite{mirza2022norm}, and diffusion-based adaptation~\cite{gao2023back} for input restoration. However, these methods are primarily designed for regular grid structures and do not readily generalize to sparse and irregular 3D point clouds.
For 3D point cloud classification, TTA is an emerging research area with distinct challenges. Early efforts in 3D TTA often rely on the Test-Time Training (TTT) paradigm~\cite{sun2020test}, where MATE~\cite{mirza2023mate} utilizes masked reconstruction and SMART-PC~\cite{bahri2025smartpc} employs skeleton learning as auxiliary self-supervised tasks during pre-training. Although these methods enhance robustness, the need for task-specific pre-training and architectural modifications limits their applicability to generic, off-the-shelf pre-trained models. In contrast, fully TTA methods operate on frozen pre-trained models. Recent fully TTA methods attempt to resolve these issues: CloudFixer~\cite{shim2024cloudfixer} and 3DD-TTA~\cite{dastmalchi2025test} use diffusion models for restoration but incur high latency, while BFTT3D~\cite{wang2024backpropagation} and PG-SP~\cite{yazdanpanah2025purge} rely on source prototypes. Although source-free variants like PG-SF exist, they often deliver limited performance gains. More recently, SVWA~\cite{bahri2025test} has achieved strong results via multi-view weight averaging. However, its reliance on sequential optimization for each view results in significant inference latency. Our approach, MAMVI, addresses these limitations by utilizing efficient masking to enable source-free adaptation on a single model through a unified backward pass.

\subsection{Masked Self-Supervised Learning}

Masked modeling has demonstrated remarkable success in learning robust representations across various modalities. BERT~\cite{devlin2019bert} pioneered masked language modeling, while MAE~\cite{he2022mae} subsequently extended this paradigm to 2D image domains. For 3D point clouds, recent approaches~\cite{pang2023masked,zhang2022point} learn robust geometric representations by reconstructing masked point patches during pre-training. These methods, however, apply masking only during pre-training and rely on complete point clouds at inference time.

MATE~\cite{mirza2023mate} utilizes masking as a self-supervised reconstruction task to align distribution shifts during adaptation. In contrast, we utilize masking to generate multiple views by masking various subsets of patches at varying ratios. Unlike reconstruction-based methods, our strategy focuses on consensus formation and normalization parameter updates during inference, eliminating the need for auxiliary pre-training tasks.

\begin{figure*}[!t]
\centering
\includegraphics[width=1.0\textwidth]{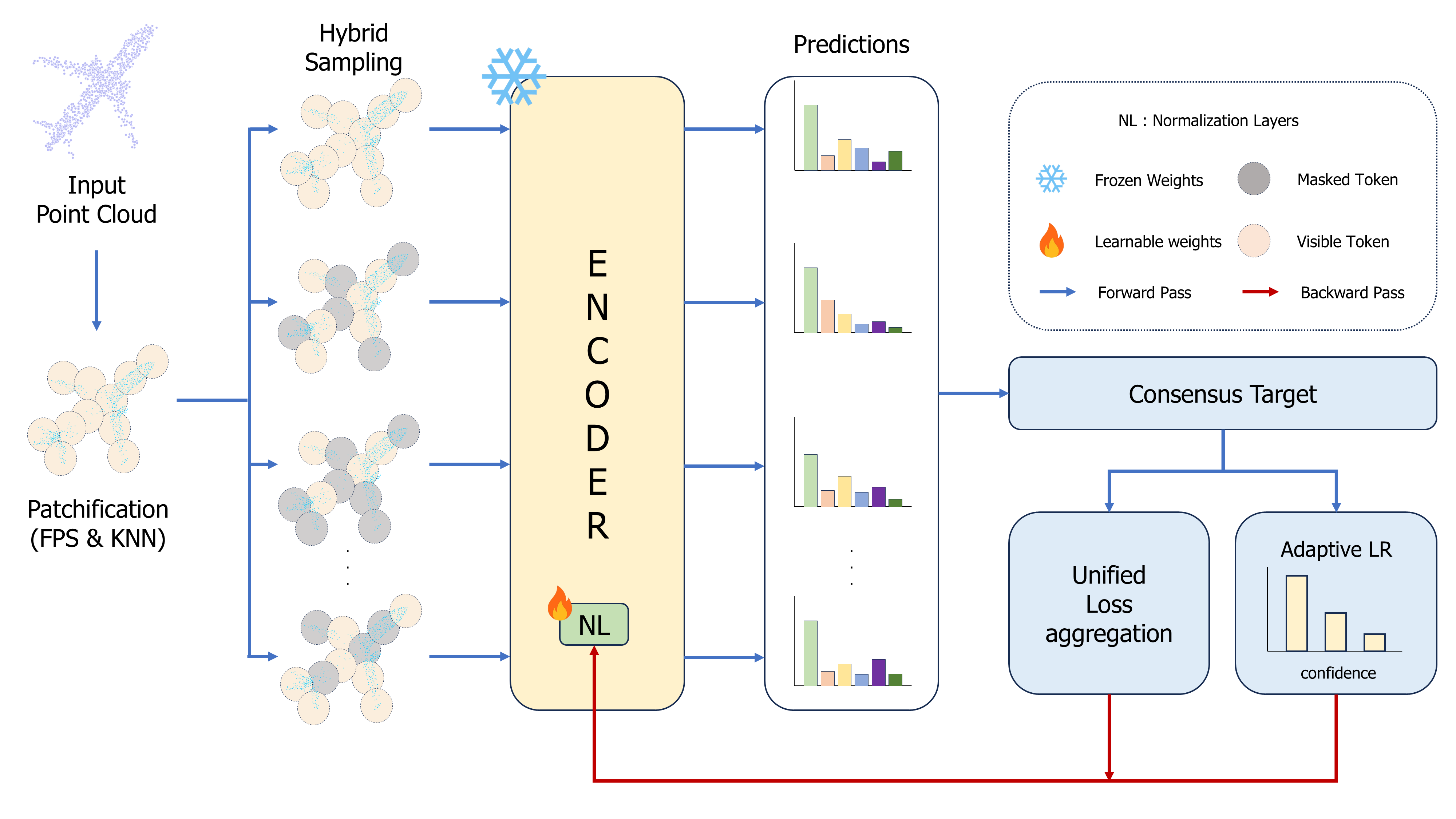}
\caption{Overview of the MAMVI framework. Masked multi-views are generated from patchified point clouds and processed through a single model. A consensus target is formed to guide unified loss aggregation and adaptive learning rate (ALR) modulation. We update only the normalization layers (NL) via a single backward pass, ensuring high inference efficiency.}
\label{fig:overview}
\end{figure*}

\section{Method}
\label{sec:method}

We propose MAMVI, a source-free test-time adaptation framework for 3D point cloud classification that enhances model robustness under distribution shifts. Our method generates various perspectives of the input data through masking at varying ratios, averages these predictions to obtain a unified target, and adapts normalization parameters using adaptive learning rates. The overview of our method is presented in \cref{fig:overview}.

\subsection{Masked Multi-View Consensus}

Given a source-trained point cloud classifier $f_\theta$ and unlabeled test samples $\{\mathbf{x}_t\}$ from a shifted target domain, we adapt the model at test time without source data access. To maintain inference efficiency, we perform a single optimization step per batch.

Formally, an input 3D point cloud is represented as $\mathbf{P} \in \mathbb{R}^{N_p \times 3}$, where $N_p$ is the number of points. To structure this irregular point cloud, we first partition it into a collection of patches. We employ Farthest Point Sampling (FPS) to select $N$ center points $\{c_i\}_{i=1}^N$, ensuring uniform geometric coverage of the overall structure. For each center, we determine its $K$ neighboring points using K-Nearest Neighbors (KNN) to group them into a patch $\mathbf{P}_i \in \mathbb{R}^{K \times 3}$. This process converts the point cloud into a set of patches $\{\mathbf{P}_i\}_{i=1}^N$.

To generate various views, we apply a hybrid masking strategy. For each input, we produce $M$ masked observations $\{\tilde{\mathbf{x}}_m\}_{m=1}^M$, where each $\tilde{\mathbf{x}}_m$ represents a partial view consisting of the patches that remain visible after masking. Specifically, we determine the masking ratios $r_m$ for these views by combining pre-defined fixed ratios with stochastic ratios sampled from a Beta distribution:
\begin{equation}
\mathcal{R} = \{r_{\text{fixed}}\} \cup \{r \sim \mathrm{Beta}(\alpha, \beta)\}.
\end{equation}
For each ratio $r_m \in \mathcal{R}$, we randomly mask $\lfloor r_m \cdot N \rfloor$ patches among the $N$ generated patches. By varying the visible patches, we obtain multiple views that allow the model to reach a more stable prediction across different masking patterns.

For each of the $M$ masked observations, the model predicts a class probability distribution:
\begin{equation}
\mathbf{p}_m = \sigma(f_\theta(\tilde{\mathbf{x}}_m)) \in \mathbb{R}^C,
\end{equation}
where $\mathbf{p}_m$ represents the predicted class probabilities for the $m$-th masked observation. We then average these predictions to form a consensus target $\bar{\mathbf{y}}$:
\begin{equation} \label{eq:consensus_dist}
\bar{\mathbf{y}} = \frac{1}{M} \sum_{m=1}^{M} \mathbf{p}_m.
\end{equation}
By averaging predictions from various masked views, we mitigate the impact of outliers. Since predictions based on partial information can be naturally unstable, aggregating multiple perspectives allows these individual inconsistencies to balance each other out. This provides a more stable and accurate target for adaptation, consistent with our experimental findings that show better performance as more views are integrated.

\subsection{Unified Loss Aggregation}

To achieve efficient single-step adaptation, we integrate the losses from all $M$ masked observations into a single optimization step. The loss for each observation incorporates both entropy minimization to increase certainty and Kullback-Leibler (KL) divergence to align individual predictions with the consensus $\bar{\mathbf{y}}$. The total loss is formulated as:
\begin{equation} \label{eq:per_view_loss}
\mathcal{L} = \frac{1}{M} \sum_{m=1}^M \left( \lambda_{\text{ent}} H(\mathbf{p}_m) + \lambda_{\text{cons}} D_{\text{KL}}(\mathbf{p}_m \| \bar{\mathbf{y}}) \right).
\end{equation}
Here, $\lambda_{\text{ent}}$ and $\lambda_{\text{cons}}$ are weighting hyperparameters that balance the two objectives. The consensus $\bar{\mathbf{y}}$ is treated as a fixed target during optimization, ensuring stable gradient flow. The entropy term encourages each view to produce confident predictions, while the KL divergence term aligns individual predictions with the collective consensus. This dual mechanism prevents the model from being misled by noisy patterns in any single masked view while maintaining overall prediction stability. By aggregating all view-specific losses into a single objective, this formulation allows the model to update its parameters through one backward pass for all $M$ views simultaneously. This eliminates the redundant optimization steps of sequential per-view methods, directly addressing the inference latency bottleneck inherent in prior multi-view adaptation methods.

\subsection{Adaptive Learning Rate Modulation}
\label{sec:alr}

To further enhance adaptation stability, we dynamically modulate the learning rate based on the consensus confidence. We first define the consensus confidence $\phi$ as the maximum probability of the aggregated multi-view prediction:
\begin{equation}
\phi = \max(\bar{\mathbf{y}}).
\end{equation}
The base learning rate $\eta_{\text{base}}$ is then modulated by a function $g(\phi)$ as follows:
\begin{equation}
\eta = \eta_{\text{base}} \times g(\phi).
\end{equation}
The modulation function $g(\phi)$ is a piecewise constant function that adjusts the adaptation step size according to the consensus confidence:
\begin{equation} \label{eq:lr_mod}
g(\phi) = \begin{cases}
m_0 & \text{if } \phi \le \tau_1, \\
m_1 & \text{if } \tau_1 < \phi \le \tau_2, \\
m_2 & \text{if } \phi > \tau_2,
\end{cases}
\end{equation}
where $\tau_1, \tau_2$ are confidence thresholds, and $m_0, m_1, m_2$ are scaling factors. This modulation function addresses a key limitation of using fixed learning rates in single-step adaptation. Test samples exhibit varying levels of adaptation difficulty, where a uniform learning rate may result in insufficient updates for complex samples or unnecessary distortion for already accurate ones. While the fixed loss coefficients $\lambda_{\text{ent}}$ and $\lambda_{\text{cons}}$ provide a static balance between the two objectives, they cannot account for the varying adaptation needs of individual samples. To address this, we dynamically modulate the learning rate based on consensus confidence. This approach directly controls the update magnitude per sample and maintains more stable gradients.

The strategy assigns higher learning rates to samples with low consensus confidence. This design is motivated by the strong correlation between consensus confidence and prediction accuracy, as demonstrated in \cref{fig:correlation_analysis}. When confidence is low, the model's prediction is likely incorrect and requires larger updates to correct its parameters. On the other hand, when confidence is high, the prediction is likely correct, and smaller updates help maintain stability without over-fitting to noise. We implement this strategy through the piecewise function $g(\phi)$, which scales the base learning rate according to confidence thresholds $\tau_1, \tau_2$ and corresponding factors $m_0, m_1, m_2$.

\section{Experiments}
\label{sec:experiments}

This section provides a comprehensive evaluation of the MAMVI method across multiple 3D point cloud classification tasks. To assess its robustness under distribution shifts, we conduct experiments on three benchmark datasets: ModelNet-40C~\cite{sun2022benchmarking}, ShapeNet-C~\cite{chang2015shapenet}, and ScanObjectNN-C~\cite{uy2019revisiting}. These datasets incorporate a range of corruption types and diverse object categories, allowing us to demonstrate the effectiveness of our approach under varied conditions.

\subsection{Implementation Details}

During adaptation, we use Point-MAE~\cite{pang2023masked} as our backbone model, initialized with pre-trained weights from MATE~\cite{mirza2023mate} to ensure a fair and consistent comparison. Following the experimental protocol of SVWA~\cite{bahri2025test}, we update only the affine parameters ($\gamma$, $\beta$) of the BatchNorm and LayerNorm layers. Furthermore, the running statistics of BatchNorm are kept frozen, focusing the adaptation exclusively on the learnable affine parameters. We conduct all experiments with a batch size of 32. We optimize MAMVI using the AdamW optimizer with a learning rate of 0.005. The configuration employs $M=12$ masked views, combining fixed ratios of $\{0, 0.2, 0.7\}$ with stochastic sampling based on a Beta distribution where $\alpha=3.0$ and $\beta=10.0$. All experiments are conducted on a single NVIDIA GeForce RTX 3090 GPU. Detailed hyperparameter configurations for both MAMVI and the compared baselines are provided in the supplementary material.

\subsection{Datasets}

\noindent\textbf{ModelNet-40C.} ModelNet-40C~\cite{sun2022benchmarking} evaluates the robustness of point cloud classification models under real-world distribution shifts. By extending the original ModelNet-40 test set with 15 distinct corruption types---spanning transformation, noise, and density perturbations---it provides a comprehensive evaluation of model performance. These corruptions emulate practical challenges such as sensor malfunctions and LiDAR noise, offering valuable insights into model behavior under authentic deployment conditions.

\noindent\textbf{ShapeNet-C.} Built upon ShapeNetCore-v2~\cite{chang2015shapenet}, a large-scale repository comprising 51,127 shapes across 55 categories, ShapeNet-C is generated by applying the 15 corruption types established in~\cite{sun2022benchmarking} to the original test set. This extension follows the standard protocol to ensure consistency across benchmarks, utilizing the open-source implementation provided by~\cite{sun2022benchmarking}.

\noindent\textbf{ScanObjectNN-C.} ScanObjectNN~\cite{uy2019revisiting} consists of 15 categories with 2,309 training samples and 581 test samples. To evaluate robustness under domain shifts, the test set is augmented with 15 corruption types following the protocol of~\cite{sun2022benchmarking}, as implemented in~\cite{mirza2023mate}. This corrupted benchmark is denoted as ScanObjectNN-C.

\subsection{Main Results}
\label{sec:main_results}

\begin{table*}[!t]
\caption{Top-1 classification accuracy (\%) for all corruption types in the ModelNet-40C dataset.}
\label{tab:modelnet}
\centering
\resizebox{\textwidth}{!}{
\small
\setlength{\tabcolsep}{2.5pt}
\begin{tabular}{l|ccccccccccccccc|c}
\toprule
Method & uni & gauss & backg & impul & upsam & rbf & rbf-inv & den-dec & dens-inc & shear & rot & cut & distort & oclsion & lidar & Mean \\
\midrule
Source Only & 66.53 & 59.12 & 7.21 & 31.69 & 74.22 & 67.71 & 69.77 & 62.28 & 75.08 & 74.35 & 38.13 & 58.63 & 70.02 & 38.53 & 18.62 & 54.13 \\
\midrule
\multicolumn{17}{c}{\textit{TTT Methods}} \\
\midrule
TTT-Rot~\cite{sun2020test} & 61.30 & 58.30 & 34.50 & 48.90 & 66.70 & 63.60 & 63.90 & 59.80 & 68.60 & 55.20 & 27.30 & 54.60 & 64.00 & 40.00 & 29.10 & 53.00 \\
MATE~\cite{mirza2023mate} & 69.80 & 61.80 & 18.90 & 63.90 & 72.50 & 64.00 & 66.00 & 74.00 & 80.80 & 71.00 & 36.70 & 69.20 & 66.30 & 38.40 & 29.90 & 58.90 \\
SMART-PC~\cite{bahri2025smartpc} & 82.40 & 80.10 & 12.00 & 67.10 & 84.50 & 76.00 & 78.60 & 67.30 & 72.90 & 73.30 & 43.90 & 72.60 & 73.50 & 37.40 & 24.80 & 63.10 \\
\midrule
\multicolumn{17}{c}{\textit{Source-Dependent Methods}} \\
\midrule
BFTT3D~\cite{wang2024backpropagation} & 66.53 & 59.20 & 7.01 & 31.85 & 65.60 & 67.79 & 69.81 & 77.96 & 85.25 & 74.35 & 38.17 & 74.80 & 70.02 & 38.61 & 30.51 & 57.16 \\
PG-SP~\cite{yazdanpanah2025purge} & 77.92 & 73.83 & 28.12 & 69.00 & 80.88 & 72.93 & 76.09 & 81.40 & 83.59 & 77.43 & 56.94 & 79.58 & 74.60 & 48.99 & 52.67 & 68.93 \\
\midrule
\multicolumn{17}{c}{\textit{Source-Free Methods}} \\
\midrule
PL~\cite{lee2013pseudo} & 74.11 & 69.61 & 19.25 & 61.71 & 68.44 & 70.10 & 70.58 & 77.55 & 81.48 & 73.50 & 53.12 & 74.68 & 70.66 & 43.35 & 39.47 & 63.17 \\
TENT~\cite{wang2020tent} & 74.96 & 71.07 & 20.26 & 61.75 & 69.57 & 70.14 & 72.08 & 77.84 & 81.89 & 73.95 & 53.12 & 75.08 & 70.34 & 44.69 & 38.01 & 63.65 \\
SHOT~\cite{liang2020we} & 62.52 & 55.83 & 19.53 & 54.74 & 58.51 & 57.58 & 62.07 & 58.91 & 69.37 & 58.75 & 45.18 & 63.78 & 58.91 & 39.38 & 37.88 & 53.52 \\
T3A~\cite{iwasawa2021test} & 65.67 & 61.40 & 21.14 & 51.66 & 67.17 & 63.27 & 65.87 & 66.36 & 69.77 & 66.36 & 50.61 & 63.43 & 62.91 & 44.89 & 43.59 & 60.24 \\
LAME~\cite{boudiaf2022parameter} & 72.16 & 59.76 & 4.34 & 33.51 & 67.26 & 67.95 & 69.94 & 78.04 & 85.82 & 74.51 & 40.32 & 74.47 & 70.91 & 37.88 & 28.32 & 57.67 \\
DUA~\cite{mirza2022norm} & 65.00 & 58.50 & 14.70 & 48.50 & 68.80 & 62.80 & 63.20 & 62.10 & 66.20 & 68.80 & 46.20 & 53.80 & 64.70 & 41.20 & 36.50 & 54.73 \\
MEMO~\cite{zhang2022memo} & 80.79 & 75.28 & 23.99 & 64.79 & 73.42 & 74.88 & 76.18 & 83.14 & 86.10 & 78.36 & 58.06 & 80.79 & 74.92 & 48.30 & 43.72 & 68.18 \\
SAR~\cite{niu2023towards} & 75.20 & 70.99 & 24.55 & 61.22 & 69.65 & 70.06 & 71.47 & 78.16 & 81.65 & 74.47 & 54.58 & 76.54 & 69.89 & 44.17 & 39.30 & 64.12 \\
PG-SF~\cite{yazdanpanah2025purge} & 78.12 & 74.49 & 40.88 & 66.69 & 81.73 & 73.58 & 75.81 & 79.09 & 82.37 & 77.23 & 58.19 & 75.49 & 73.58 & 47.85 & 45.14 & 68.68 \\
SVWA~\cite{bahri2025test} & 83.64 & \textbf{83.97} & 33.08 & \textbf{77.39} & \textbf{86.89} & \textbf{80.40} & \textbf{82.71} & \textbf{84.78} & \textbf{87.95} & 82.39 & \textbf{67.33} & \textbf{83.73} & \textbf{80.36} & \textbf{55.80} & 55.11 & \textbf{75.04} \\
\rowcolor{gray!20}\textbf{MAMVI (Ours)} & \textbf{84.01} & 82.35 & \textbf{41.48} & 73.50 & 85.84 & 79.42 & 81.98 & 84.01 & 86.40 & \textbf{82.47} & 66.27 & 83.12 & 79.99 & 53.61 & \textbf{57.55} & 74.80 \\
\bottomrule
\end{tabular}
}
\end{table*}

\begin{table*}[!t]
\caption{Top-1 classification accuracy (\%) for all corruption types in the ShapeNet-C dataset.}
\label{tab:shapenet}
\centering
\resizebox{\textwidth}{!}{
\small
\setlength{\tabcolsep}{2.5pt}
\begin{tabular}{l|ccccccccccccccc|c}
\toprule
Method & uni & gauss & backg & impul & upsam & rbf & rbf-inv & den-dec & dens-inc & shear & rot & cut & distort & oclsion & lidar & Mean \\
\midrule
Source Only & 77.34 & 71.78 & 8.64 & 54.51 & 77.90 & 75.50 & 76.06 & 85.25 & 76.42 & 80.48 & 57.08 & \textbf{85.12} & 76.05 & 10.97 & 7.13 & 61.35 \\
\midrule
\multicolumn{17}{c}{\textit{TTT Methods}} \\
\midrule
TTT-Rot~\cite{sun2020test} & 74.60 & 72.40 & 23.10 & 59.90 & 74.90 & 73.80 & 75.00 & 81.40 & 82.00 & 69.20 & 49.10 & 79.90 & 72.70 & \textbf{14.00} & 12.00 & 60.90 \\
MATE~\cite{mirza2023mate} & 77.80 & 74.70 & 4.30 & 66.20 & 78.60 & 76.30 & 75.30 & \textbf{86.10} & \textbf{86.60} & 79.20 & 56.10 & 84.10 & 76.10 & 12.30 & 13.10 & 63.10 \\
SMART-PC~\cite{bahri2025smartpc} & 80.80 & 78.90 & 8.90 & 60.40 & \textbf{81.80} & \textbf{81.10} & \textbf{81.70} & 84.80 & 78.40 & 80.80 & 63.70 & 84.90 & \textbf{79.80} & 11.50 & 8.80 & 64.40 \\
\midrule
\multicolumn{17}{c}{\textit{Source-Dependent Methods}} \\
\midrule
BFTT3D~\cite{wang2024backpropagation} & 72.58 & 65.91 & 10.57 & 60.86 & 67.78 & 73.45 & 74.38 & 84.55 & 83.52 & 78.58 & 57.96 & 83.71 & 75.17 & 11.95 & 10.00 & 60.73 \\
PG-SP~\cite{yazdanpanah2025purge} & 78.63 & 76.52 & 15.78 & \textbf{72.70} & 78.78 & 76.49 & 77.18 & 83.28 & 81.18 & 79.05 & 63.36 & 83.07 & 76.33 & 9.85 & 10.34 & 64.17 \\
\midrule
\multicolumn{17}{c}{\textit{Source-Free Methods}} \\
\midrule
PL~\cite{lee2013pseudo} & 70.30 & 64.20 & 7.86 & 58.00 & 64.22 & 72.25 & 71.61 & 79.92 & 79.16 & 74.90 & 59.25 & 78.30 & 72.19 & 10.30 & 10.50 & 59.41 \\
TENT~\cite{wang2020tent} & 71.90 & 67.48 & 9.29 & 63.53 & 66.87 & 72.55 & 73.48 & 82.09 & 81.35 & 76.56 & 59.82 & 81.03 & 72.65 & 8.52 & 8.93 & 61.01 \\
SHOT~\cite{liang2020we} & 27.78 & 21.97 & 3.26 & 22.81 & 23.24 & 24.27 & 31.22 & 33.86 & 31.00 & 27.86 & 22.57 & 34.19 & 28.73 & 3.48 & 2.77 & 22.84 \\
T3A~\cite{iwasawa2021test} & 66.69 & 64.18 & 14.00 & 59.40 & 67.14 & 64.32 & 64.90 & 71.07 & 66.69 & 66.34 & 54.76 & 69.42 & 64.15 & 9.89 & 10.46 & 54.23 \\
LAME~\cite{boudiaf2022parameter} & 72.88 & 66.90 & 4.90 & 58.43 & 64.48 & 73.54 & 74.63 & 84.40 & 83.58 & 78.45 & 53.44 & 83.36 & 75.25 & 10.39 & 8.48 & 60.85 \\
DUA~\cite{mirza2022norm} & 76.10 & 70.10 & 14.30 & 60.90 & 76.20 & 71.60 & 72.90 & 80.00 & 83.80 & 77.10 & 57.50 & 75.00 & 72.10 & 11.90 & 12.10 & 60.77 \\
MEMO~\cite{zhang2022memo} & 75.27 & 71.58 & 12.47 & 68.03 & 70.86 & 75.88 & 76.76 & 84.40 & 83.70 & 79.98 & 65.63 & 83.53 & 76.10 & 10.63 & 11.49 & 64.30 \\
SAR~\cite{niu2023towards} & 72.49 & 67.68 & 9.65 & 64.26 & 67.12 & 73.14 & 73.84 & 82.44 & 81.62 & 77.54 & 59.56 & 81.60 & 72.55 & 9.58 & 10.52 & 61.46 \\
PG-SF~\cite{yazdanpanah2025purge} & 79.21 & 75.43 & 22.46 & 68.38 & 78.95 & 76.78 & 77.12 & 83.52 & 80.73 & 79.24 & 64.91 & 83.55 & 76.08 & 10.15 & 9.55 & 64.40 \\
SVWA~\cite{bahri2025test} & \textbf{82.04} & \textbf{80.31} & \textbf{24.93} & 70.98 & 81.39 & 78.84 & 79.19 & 82.84 & 81.40 & 78.11 & 67.86 & 80.07 & 75.06 & 11.35 & 11.79 & 65.74 \\
\rowcolor{gray!20}\textbf{MAMVI (Ours)} & 81.70 & 79.11 & 21.54 & 67.25 & 81.42 & 79.72 & 80.26 & 84.96 & 83.98 & \textbf{81.27} & \textbf{69.17} & 84.59 & 78.75 & 12.94 & \textbf{14.94} & \textbf{66.77} \\
\bottomrule
\end{tabular}
}
\end{table*}

\begin{table*}[!t]
\caption{Top-1 classification accuracy (\%) for all corruption types in the ScanObjectNN-C dataset.}
\label{tab:scanobj}
\centering
\resizebox{\textwidth}{!}{
\small
\setlength{\tabcolsep}{2.5pt}
\begin{tabular}{l|ccccccccccccccc|c}
\toprule
Method & uni & gauss & backg & impul & upsam & rbf & rbf-inv & den-dec & dens-inc & shear & rot & cut & distort & oclsion & lidar & Mean \\
\midrule
Source Only & 21.69 & 18.76 & 16.70 & 18.42 & 22.20 & 45.96 & 46.99 & 71.95 & 69.36 & 48.88 & 35.63 & 72.98 & 49.40 & 6.71 & 8.78 & 36.96 \\
\midrule
\multicolumn{17}{c}{\textit{TTT Methods}} \\
\midrule
TTT-Rot~\cite{sun2020test} & 30.20 & 34.10 & 16.20 & 22.60 & 22.60 & 32.40 & 32.10 & 45.50 & 45.00 & 34.50 & 29.30 & 47.80 & 36.20 & 7.10 & 8.10 & 29.58 \\
MATE~\cite{mirza2023mate} & 27.50 & 29.40 & 14.30 & 22.20 & 25.60 & 40.80 & 42.00 & 73.70 & 63.20 & 45.10 & 35.30 & 73.30 & 45.30 & 7.10 & 9.30 & 36.90 \\
SMART-PC~\cite{bahri2025smartpc} & 27.50 & 39.10 & 19.30 & 21.50 & 29.80 & 44.20 & 48.90 & 68.30 & 60.20 & 49.40 & 45.40 & 70.10 & 49.10 & 8.40 & \textbf{12.20} & 39.60 \\
\midrule
\multicolumn{17}{c}{\textit{Source-Dependent Methods}} \\
\midrule
BFTT3D~\cite{wang2024backpropagation} & 16.35 & 14.97 & 14.29 & 17.04 & 13.43 & 38.90 & 42.00 & 69.71 & 70.74 & 41.14 & 28.92 & 69.19 & 43.20 & 6.54 & 8.61 & 33.00 \\
PG-SP~\cite{yazdanpanah2025purge} & 41.48 & 55.59 & 22.89 & 38.90 & 42.17 & 56.80 & 62.31 & \textbf{76.08} & \textbf{76.42} & 59.04 & 49.74 & \textbf{76.76} & 62.48 & 7.57 & 8.78 & 49.13 \\
\midrule
\multicolumn{17}{c}{\textit{Source-Free Methods}} \\
\midrule
PL~\cite{lee2013pseudo} & 39.76 & 36.83 & 19.10 & 43.55 & 37.87 & 57.14 & 62.31 & 66.27 & 66.27 & \textbf{60.59} & 49.74 & 67.30 & 61.79 & 8.26 & 9.12 & 46.51 \\
TENT~\cite{wang2020tent} & 39.24 & 34.42 & 17.04 & 44.75 & 34.77 & \textbf{58.00} & 59.04 & 66.95 & 66.95 & 60.24 & 50.26 & 67.64 & 61.27 & 7.06 & 9.12 & 45.97 \\
SHOT~\cite{liang2020we} & 25.82 & 34.25 & 13.08 & \textbf{54.04} & 32.53 & 57.49 & 58.86 & 66.44 & 66.61 & 54.73 & 46.99 & 65.92 & 63.34 & 6.02 & 8.26 & 44.57 \\
T3A~\cite{iwasawa2021test} & 37.50 & 46.01 & 18.92 & 37.33 & 37.67 & 50.69 & 52.26 & 63.72 & 65.97 & 50.35 & 46.88 & 64.41 & 54.17 & \textbf{9.72} & 8.68 & 42.95 \\
LAME~\cite{boudiaf2022parameter} & 14.46 & 12.56 & 10.33 & 15.83 & 12.91 & 48.88 & 55.59 & 58.35 & 61.10 & 52.32 & 35.80 & 58.35 & 53.36 & 9.12 & 9.47 & 34.91 \\
DUA~\cite{mirza2022norm} & 30.50 & 40.10 & 10.20 & 23.60 & 29.90 & 43.70 & 46.10 & 68.30 & 66.30 & 48.50 & 38.90 & 68.70 & 48.40 & 8.60 & 8.10 & 38.66 \\
MEMO~\cite{zhang2022memo} & - & - & - & - & - & - & - & - & - & - & - & - & - & - & - & - \\
SAR~\cite{niu2023towards} & 41.65 & 38.73 & 18.24 & 46.47 & 39.24 & 57.14 & 62.65 & 66.61 & 65.58 & 60.41 & \textbf{52.32} & 68.16 & 62.48 & 8.95 & 9.47 & 47.26 \\
PG-SF~\cite{yazdanpanah2025purge} & 40.96 & 54.76 & 21.86 & 36.49 & 39.59 & 55.25 & 58.35 & 75.56 & 74.04 & 56.80 & 48.71 & 75.56 & 58.69 & 7.06 & 9.47 & 47.54 \\
SVWA~\cite{bahri2025test} & \textbf{44.62} & \textbf{60.94} & \textbf{25.17} & 38.72 & \textbf{46.70} & 52.60 & 55.56 & 72.05 & 72.40 & 52.60 & 44.44 & 71.18 & 53.99 & 9.38 & 11.46 & 47.45 \\
\rowcolor{gray!20}\textbf{MAMVI (Ours)} & 43.06 & 60.07 & 22.92 & 41.15 & 45.49 & 56.77 & 61.81 & 74.83 & 74.83 & 56.77 & 48.79 & \textbf{77.26} & \textbf{63.72} & 8.33 & 10.59 & \textbf{49.76} \\
\bottomrule

\end{tabular}
}
\end{table*}

\noindent\textbf{ModelNet-40C.} 
As shown in \Cref{tab:modelnet}, MAMVI achieves a mean accuracy of 74.80\% on ModelNet-40C, representing a substantial improvement of +20.67\% over the Source Only baseline. It significantly outperforms conventional TTA methods such as TENT (63.65\%) and SHOT (53.52\%), as well as recent 3D-specific approaches including Purge-Gate (PG-SP: 68.93\%; PG-SF: 68.68\%). While SVWA achieves a slightly higher mean accuracy of 75.04\%, MAMVI bridges the trade-off between accuracy and inference efficiency with an 8.6$\times$ speedup as shown in \Cref{tab:efficiency}. Furthermore, MAMVI achieves the best performance on several individual corruption types, such as \textit{uniform} and \textit{lidar}.

\noindent\textbf{ShapeNet-C.}
\Cref{tab:shapenet} demonstrates that on the ShapeNet-C dataset, MAMVI achieves a state-of-the-art mean accuracy of 66.77\%, outperforming all compared adaptation methods. It shows consistent gains over existing benchmarks, including TENT (61.01\%) and MEMO (64.30\%), and notably surpasses recent 3D-specific methods such as PG-SP (64.17\%), PG-SF (64.40\%), and SVWA (65.74\%). Notably, MAMVI achieves the best performance on several challenging corruptions, including \textit{shear} and \textit{lidar}, demonstrating its effectiveness in handling complex environmental noise and sensor variations.

\noindent\textbf{ScanObjectNN-C.}
\Cref{tab:scanobj} highlights MAMVI's strong performance on the real-world ScanObjectNN-C benchmark, achieving a state-of-the-art mean accuracy of 49.76\%, representing a +12.80\% improvement over the Source Only baseline. It outperforms conventional methods such as TENT (45.97\%) and SHOT (44.57\%), and demonstrates consistent gains over recent 3D-specific TTA approaches including PG-SP (49.13\%), PG-SF (47.54\%), and SVWA (47.45\%). These results demonstrate MAMVI's effectiveness across diverse real-world scenarios.

\begin{table}[!t]
\centering
\small
\caption{Inference throughput (Frames/Second) comparison between MAMVI and SVWA across datasets.}
\label{tab:efficiency}
\begin{tabular}{l|ccc}
\toprule
\textbf{Method} & \textbf{ModelNet-40C} & \textbf{ShapeNet-C} & \textbf{ScanObjectNN-C} \\
\midrule
SVWA (Frames/Second) & 8.99 & 8.63 & 9.03 \\
\rowcolor{gray!20}\textbf{MAMVI (Frames/Second)} & \textbf{77.10} & \textbf{76.64} & \textbf{44.50} \\
\midrule
\textbf{Speedup} & \textbf{8.6$\times$} & \textbf{8.9$\times$} & \textbf{4.9$\times$} \\
\bottomrule
\end{tabular}
\end{table}

\begin{figure}[!t]
\centering
\includegraphics[width=1.0\columnwidth]{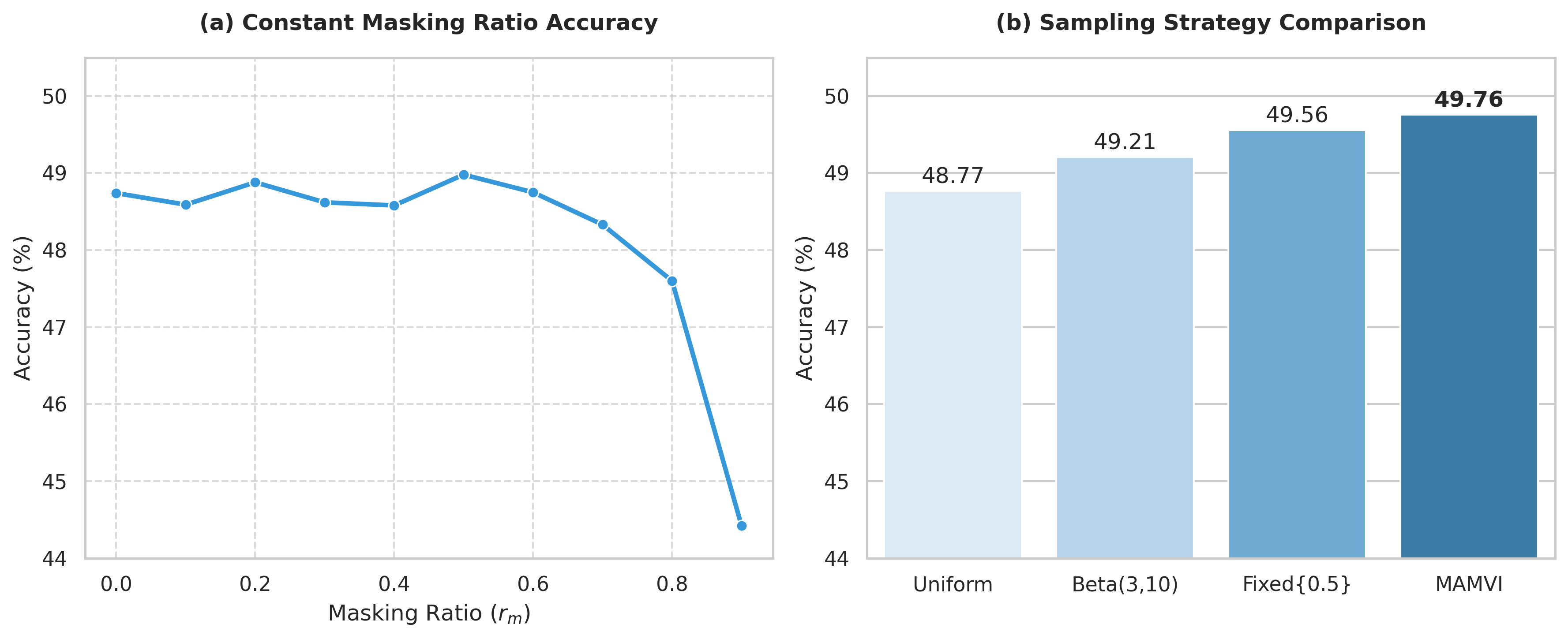}
\caption{Analysis of masking strategies on ScanObjectNN-C. (a) Classification accuracy across varying constant masking ratios. (b) Comparison of different sampling strategies, where MAMVI yields the best performance.}
\label{fig:masking_ablation}
\end{figure}

\subsection{Ablation Study and Analysis}
\subsubsection{Real-Time Adaptation.}
\Cref{tab:efficiency} illustrates the inference throughput of MAMVI and SVWA in terms of Frames per Second (FPS) across ModelNet-40C, ShapeNet-C, and ScanObjectNN-C. Both methods are evaluated under identical experimental conditions to ensure a fair comparison. MAMVI consistently demonstrates exceptional efficiency, delivering significant speedups of 8.6$\times$, 8.9$\times$, and 4.9$\times$ on ModelNet-40C, ShapeNet-C, and ScanObjectNN-C, respectively. This acceleration is achieved by replacing the costly sequential optimization of prior methods with our unified single-step approach. By aggregating losses from multiple masked views into a single optimization objective, MAMVI eliminates redundant optimization iterations and resolves the inference latency bottleneck inherent in prior multi-view adaptation frameworks.

\subsubsection{Masking Strategy Ablation.}

The results in \Cref{fig:masking_ablation} on ScanObjectNN-C reveal several key insights. As shown in (a), using a constant masking ratio across all $M$ views leads to diminished performance due to a lack of view diversity. In (b), while solely utilizing the $\mathrm{Beta}(3, 10)$ distribution provides better diversity than uniform sampling, its effectiveness is further enhanced when combined with fixed anchors in a hybrid configuration. Notably, incorporating an unmasked view ($r_m=0.0$) with proper masking ratios (e.g., $\{0.0, 0.2, 0.7\}$) achieves the highest performance compared to other hybrid setups such as $\{0.5, 0.5, 0.5\}$. These findings validate that our hybrid sampling strategy, which leverages the complementary strengths of the stochastic Beta distribution and carefully selected fixed ratios, provides a more robust and effective masking approach for test-time adaptation.

\begin{figure*}[!t]
\centering
\begin{minipage}[b]{0.48\textwidth}
    \centering
    \includegraphics[width=\textwidth,height=5.5cm,keepaspectratio]{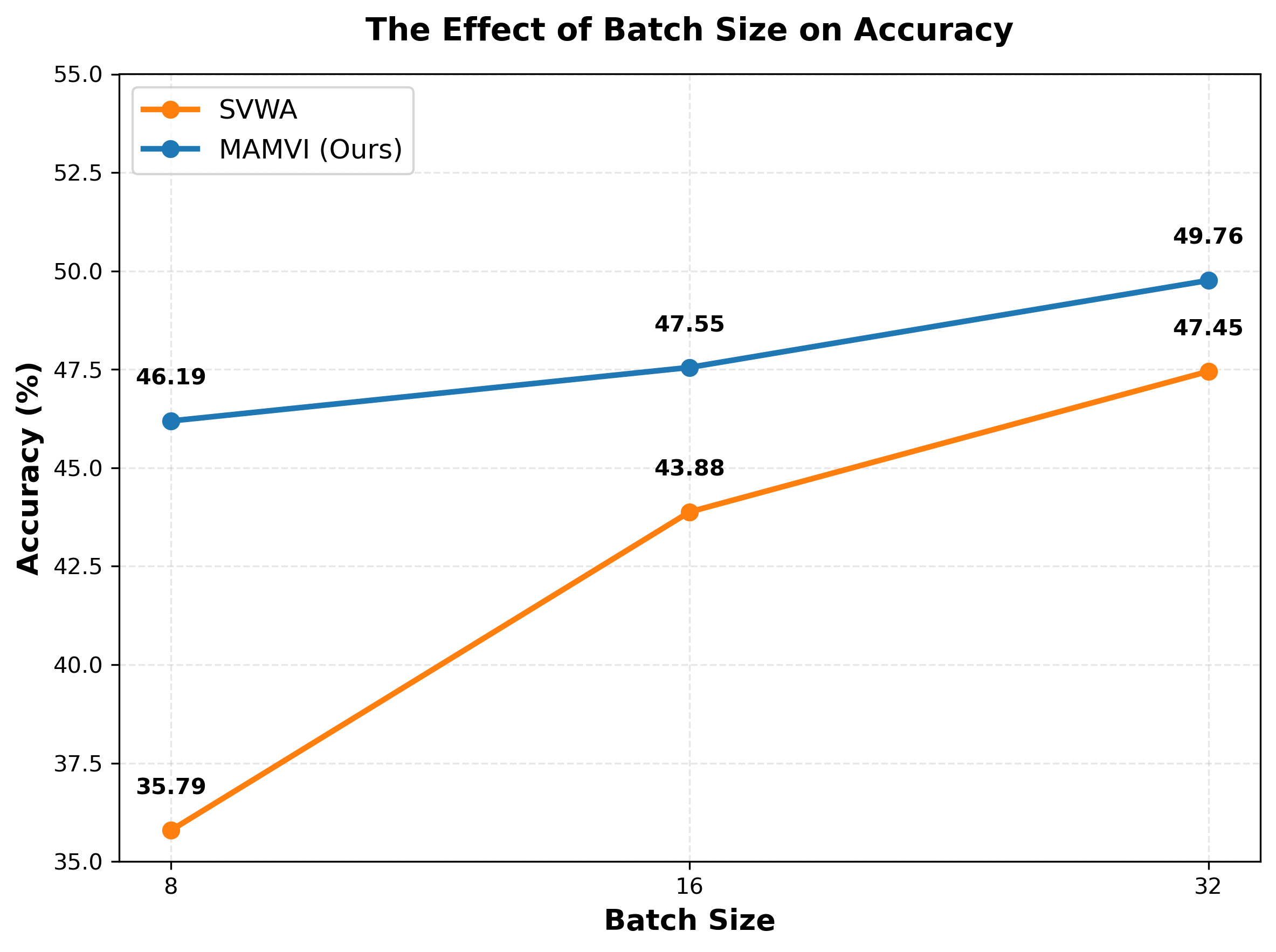}
    \caption{Impact of batch size on accuracy for two methods. The experiments were conducted on ScanObjectNN-C.}
    \label{fig:batch_size_sensitivity}
\end{minipage}
\hfill
\begin{minipage}[b]{0.48\textwidth}
    \centering
    \includegraphics[width=\textwidth,height=5.5cm,keepaspectratio]{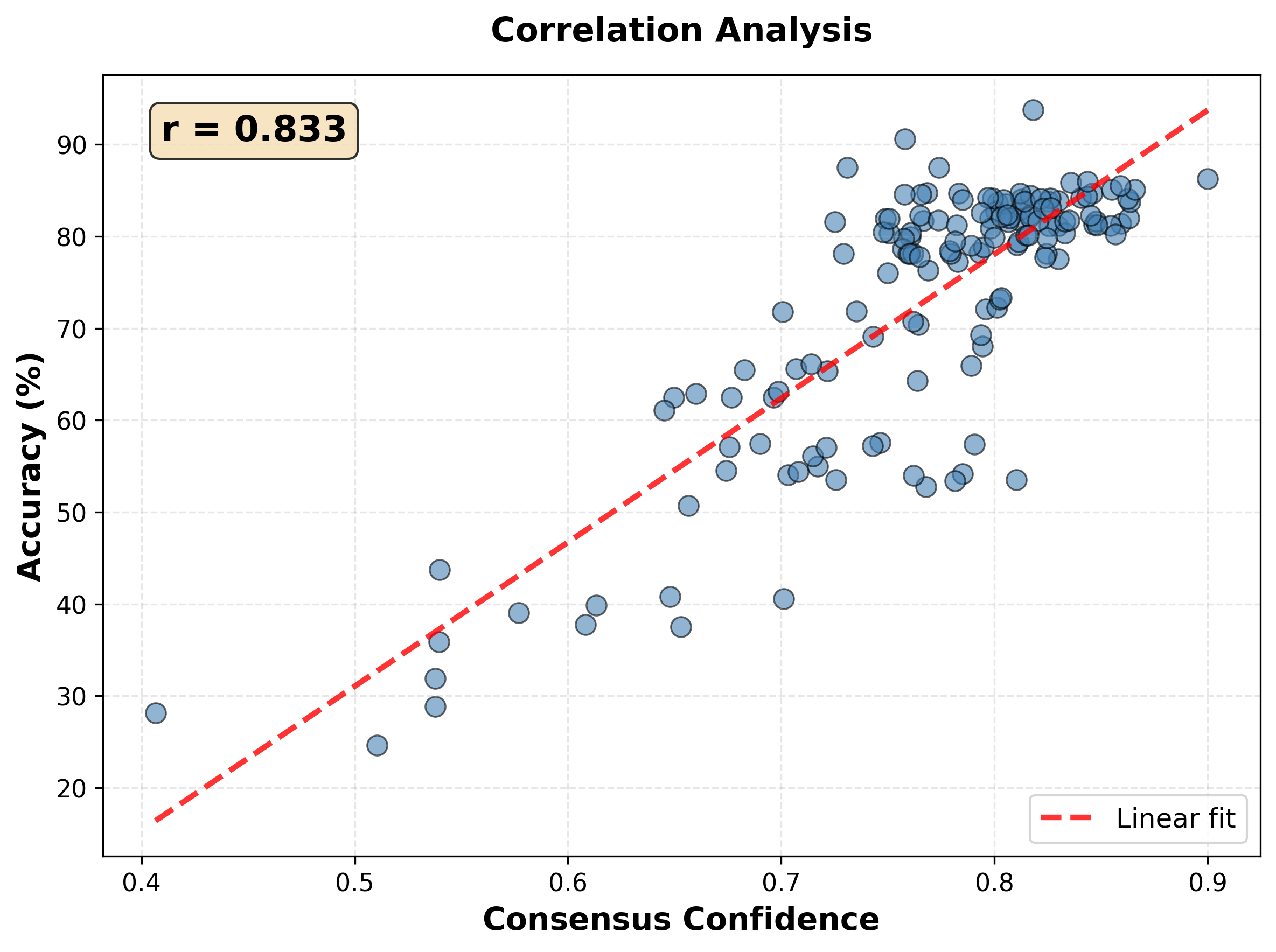}
    \caption{Correlation between consensus confidence and accuracy. The analysis was conducted on ModelNet-40C.}
    \label{fig:correlation_analysis}
\end{minipage}
\end{figure*}
\subsubsection{Batch Size Sensitivity.}
\Cref{fig:batch_size_sensitivity} compares batch size sensitivity of MAMVI against SVWA on ScanObjectNN-C. MAMVI achieves 46.19\% accuracy at batch size 8, 47.55\% at batch size 16, and 49.76\% at batch size 32, while SVWA obtains 35.79\%, 43.88\%, and 47.45\% respectively. The results show that MAMVI maintains more stable performance across different batch sizes, with a performance gap of +10.40\% at batch size 8 that narrows to +2.31\% at batch size 32. This demonstrates that our approach is robust across different batch settings, making it effective even in scenarios with limited data.

\subsubsection{Correlation Analysis.}
\Cref{fig:correlation_analysis} validates the key assumption underlying our Adaptive Learning Rate mechanism (\cref{sec:alr}). We analyze the relationship between consensus confidence and prediction accuracy across 150 intermediate checkpoints during adaptation on ModelNet-40C. The results show a strong positive Pearson correlation ($r=0.833$), confirming that higher consensus confidence reliably indicates correct predictions. This empirical evidence justifies our strategy of assigning lower learning rates to high-confidence samples (which are likely correct and require stability) and higher learning rates to low-confidence samples (which are likely incorrect and require larger corrective updates).

\subsubsection{Adaptive Learning Rate Ablation.}

\begin{table*}[!t]
\centering
\small
\caption{Impact of the adaptive learning rate (ALR) mechanism across benchmarks.}
\label{tab:adaptive_lr}
\begin{tabular*}{\textwidth}{@{\extracolsep{\fill}}lccc}
\toprule
Method & ModelNet-40C & ShapeNet-C & ScanObjectNN-C \\
\midrule
No ALR & 73.86 & 65.63 & 49.33 \\
MAMVI (ALR) & \textbf{74.80} & \textbf{66.77} & \textbf{49.76} \\
\midrule
Gain ($\Delta$) & +0.94 & +1.14 & +0.43 \\
\bottomrule
\end{tabular*}
\end{table*}

\Cref{tab:adaptive_lr} evaluates the impact of the adaptive learning rate (ALR) mechanism. Incorporating ALR consistently yields performance gains, improving mean accuracy by +0.94\% on ModelNet-40C, +1.14\% on ShapeNet-C, and +0.43\% on ScanObjectNN-C. These gains across all three benchmarks demonstrate that the confidence-based modulation effectively balances adaptation aggressiveness and stability within a single optimization step.

\begin{figure}[!t]
\centering
\includegraphics[width=1.0\columnwidth]{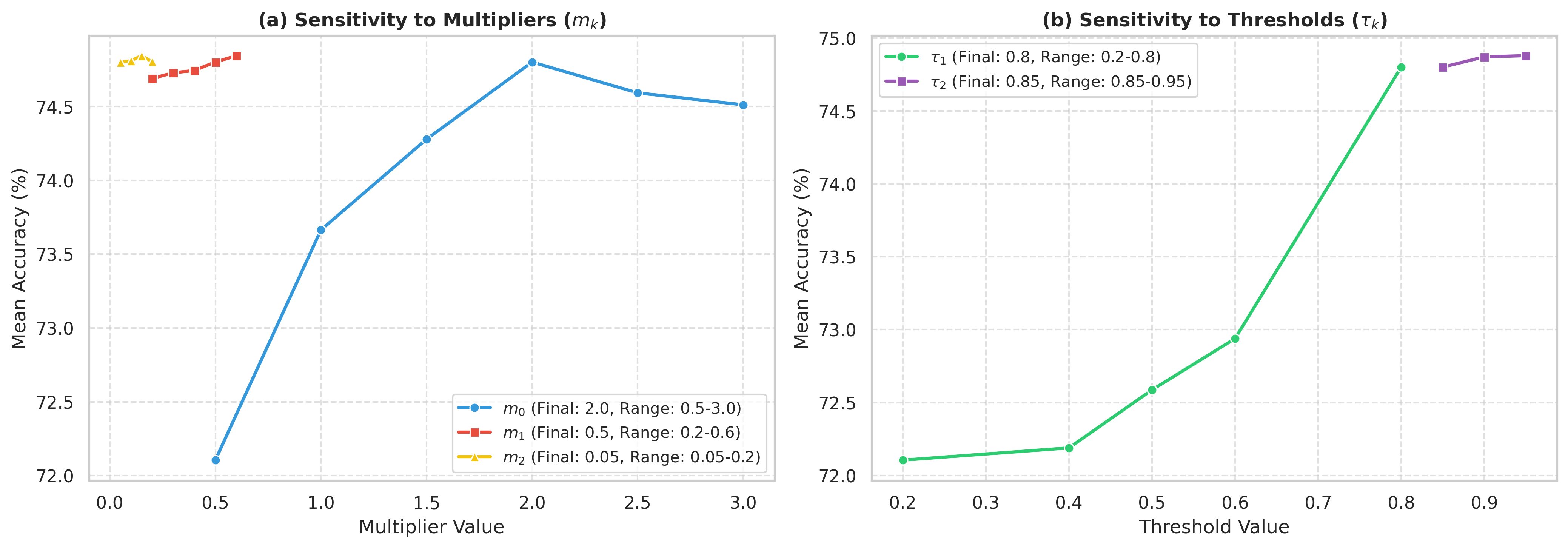}
\caption{Sensitivity analysis of individual Adaptive Learning Rate (ALR) parameters on ModelNet-40C. We evaluate (a) Multipliers ($m_0, m_1, m_2$) and (b) Thresholds ($\tau_1, \tau_2$). Values in legend denote the sweep range for each parameter.}
\label{fig:alr_sensitivity}
\end{figure}

\Cref{fig:alr_sensitivity} further analyzes the sensitivity of the Adaptive Learning Rate (ALR) module to its internal parameters on ModelNet-40C. As shown in (a), the uncertainty multiplier $m_0$ significantly impacts performance, with accuracy improving as $m_0$ increases from 0.5 to 3.0. This confirms that providing aggressive corrective updates for low-confidence samples is the primary driver of adaptation performance. In contrast, the relatively flat lines for $m_1$ (0.2--0.6) and $m_2$ (0.05--0.2) demonstrate the robustness of MAMVI across a wide range of multipliers. In (b), the threshold $\tau_1$ shows a clear peak at 0.8 within its sweep range (0.2--0.8), validating it as a reliable boundary for identifying samples requiring correction, while the minimal variation across $\tau_2$ (0.85--0.95) reflects stable handling of high-confidence predictions. Based on these observations, we fix the final hyperparameters as $m_0=2.0$, $m_1=0.5$, $m_2=0.05$, $\tau_1=0.8$, and $\tau_2=0.85$ for all benchmarks.

\subsubsection{Hyperparameter Ablation.}

\begin{figure*}[!t]
\centering
\includegraphics[width=1.0\textwidth]{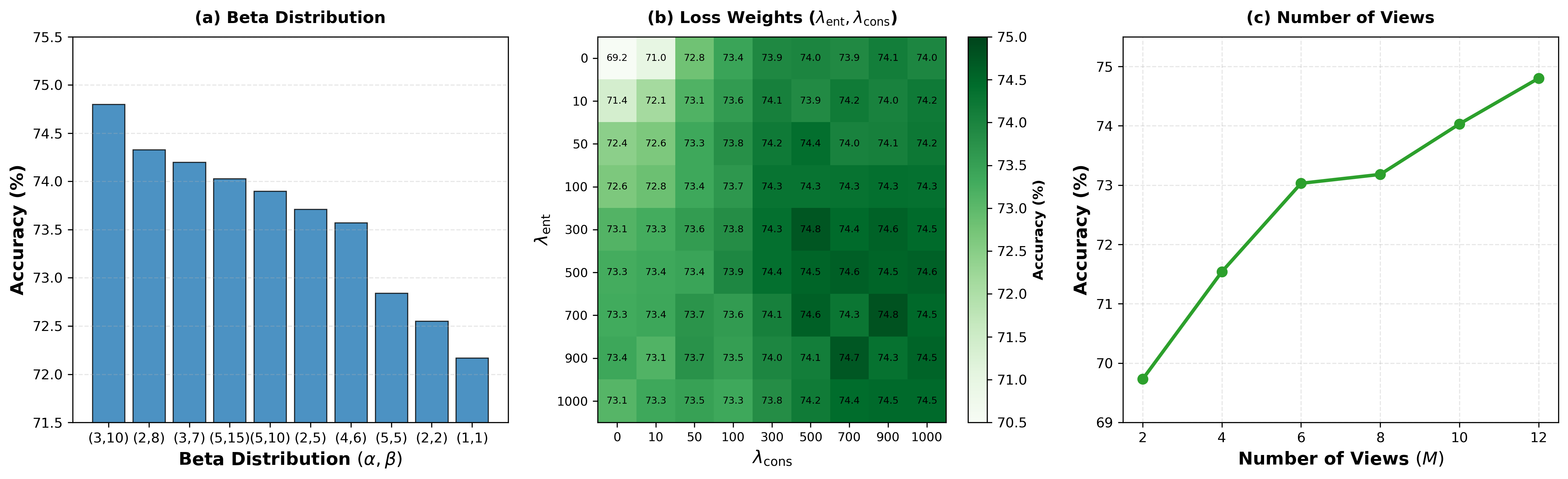}
\caption{Comprehensive hyperparameter ablation studies on ModelNet-40C. We investigate (a) Beta distribution parameters $(\alpha, \beta)$, (b) Loss weights for entropy and consensus consistency ($\lambda_{\text{ent}}$, $\lambda_{\text{cons}}$), and (c) Impact of the number of views ($M$).}
\label{fig:hyperparameter_ablation}
\end{figure*}

\Cref{fig:hyperparameter_ablation} presents comprehensive hyperparameter ablation studies on ModelNet-40C. As shown in (a), the $\mathrm{Beta}(3, 10)$ distribution (mean ratio $\approx$ 0.23) proves most effective among various Beta configurations. While more uniform distributions like $\mathrm{Beta}(1, 1)$ or $\mathrm{Beta}(2, 2)$ introduce excessive uncertainty through heavy masking, $\mathrm{Beta}(3, 10)$ maintains the optimal balance between sufficient view diversity and conservation of geometric details.
In (b), we evaluate the synergy between entropy minimization and consensus consistency. Without adaptation, the model achieves only 69.22\%. While using either loss term alone provides moderate improvements, using both terms with balanced weights ($\lambda_{\text{ent}}=700, \lambda_{\text{cons}}=900$) improves performance by up to +1.42\% compared to the strongest single-loss setting. This confirms that the two objectives are complementary and together yield a significant synergistic effect.
Finally, (c) illustrates that accuracy increases consistently as the number of masked views ($M$) grows, starting from 69.73\% at $M=2$. Since MAMVI requires only a single backward pass regardless of the number of views, we select $M=12$ as our default. This configuration ensures both maximum adaptation stability and high inference efficiency, overcoming the linear computational overhead typical of prior multi-view methods.

\section{Conclusion}

In this paper, we introduced MAMVI, a novel masked multi-view framework for single-step test-time adaptation of 3D point cloud models. By integrating a hybrid masking strategy with adaptive learning rate modulation, MAMVI achieves highly stable and effective adaptation. Our approach effectively balances view diversity and consensus stability, enabling a single aggregated update that captures robust geometric features under domain shifts. Extensive evaluations on ModelNet-40C, ShapeNet-C, and ScanObjectNN-C demonstrate that MAMVI achieves state-of-the-art performance on ShapeNet-C and ScanObjectNN-C, while maintaining competitive results on ModelNet-40C. Ultimately, MAMVI provides a robust and reliable solution for 3D perception in dynamic environments.

\section*{Acknowledgments and Disclosure of Funding}
This research was supported by the MSIT(Ministry of Science and ICT), Korea, under the National Program for Excellence in SW supervised by the IITP(Institute of Information \& Communications Technology Planning \& Evaluation) in 2026. Also, this research was supported by the National Research Foundation of Korea(NRF) grant funded by the korea government(MSIT) for Jiyoung Jung (RS-2025-24523036).

%
%
%
\bibliographystyle{splncs04}
\bibliography{Citation}

\end{document}